\documentclass[10pt,twocolumn,letterpaper]{article}

\usepackage{cvpr}
\usepackage{times}
\usepackage{epsfig}
\usepackage{graphicx}
\usepackage{amsmath}
\usepackage{amssymb}
\usepackage{algorithm}
\usepackage{algorithmic}
\usepackage{color}
\usepackage{bm}
\usepackage{booktabs}
\DeclareMathOperator*{\argmin}{argmin}

\usepackage[breaklinks=true,bookmarks=false]{hyperref}

\cvprfinalcopy 

\def\cvprPaperID{3439} 

\begin{document}

\title{A Resizable Mini-batch Gradient Descent based on a Multi-Armed Bandit}

\author{Seong Jin Cho \qquad \qquad Sunghun Kang \qquad \qquad Chang D. Yoo \\
	Korea Advanced Institute of Science and Technology (KAIST)\\
	291 Daehak-ro, Yuseong-gu, Daejeon 34141, Republic of Korea\\
	{\tt\small \{ipcng00, sunghun.kang, cd\_yoo\} @kaist.ac.kr}
}

\maketitle

\begin{abstract}
	Determining the appropriate batch size for mini-batch gradient descent is always time consuming as it often relies on grid search. This paper considers a resizable mini-batch gradient descent (RMGD) algorithm based on a multi-armed bandit for achieving best performance in grid search by selecting an appropriate batch size at each epoch with a probability defined as a function of its previous success/failure. This probability encourages exploration of different batch size and then later exploitation of batch size with history of success. At each epoch, the RMGD samples a batch size from its probability distribution, then uses the selected batch size for mini-batch gradient descent. After obtaining the validation loss at each epoch, the probability distribution is updated to incorporate the effectiveness of the sampled batch size. The RMGD essentially assists the learning process to explore the possible domain of the batch size and exploit successful batch size. Experimental results show that the RMGD achieves performance better than the best performing single batch size. Furthermore, it, obviously, attains this performance in a shorter amount of time than grid search. It is surprising that the RMGD achieves better performance than grid search.
\end{abstract}

\section{Introduction}

Gradient descent (GD) is a common algorithm in minimizing the expected loss. It takes iterative steps proportional to the negative gradient of the loss function at each iteration. It is based on the observation that if the multi-variable loss functions $f(\bm{w})$ is differentiable at a point $\bm{w}$, then $f(\bm{w})$ decreases fastest in the direction of the negative gradient of $f$ at $\bm{w}$, i.e., $-\nabla f(\bm{w})$. The model parameters are updated iteratively in GD as follows:
\begin{equation*}
\bm{w}_{t+1} = \bm{w}_t - \eta_t \bm{g}_t, \qquad \bm{g}_t = \nabla_{\bm{w}} f (\bm{w}_t)
\end{equation*}

\noindent where $\bm{w}_t$, $\bm{g}_t$, and $\eta_t$ are the model parameters, gradients of $f$ with respect to $\bm{w}$, and learning rate at time $t$ respectively. For small enough $\eta_t$, $f(\bm{w}_t) \ge f(\bm{w}_{t+1})$ and ultimately the sequence of $\bm{w}_t$ will move down toward a local minimum. For a convex loss function, GD is guaranteed to converge to a global minimum with an appropriate learning rate.

There are various issues to consider in gradient-based optimization. First, GD can be extremely slow and impractical for large dataset: gradients of all the data have to be evaluated for each iteration. With larger data size, the convergence rate, the computational cost and memory become critical, and special care is required to minimize these factors. Second, for non-convex function which is often encountered in deep learning, GD can get stuck in a local minimum without the hope of escaping. Third, stochastic gradient descent (SGD), which is based on the gradient of a single training sample, has large gradient variance, and it requires a large number of iterations. This ultimately translates to slow convergence. Mini-batch gradient descent (MGD), which is based on the gradient over a small batch of training data, trades off between the robustness of SGD and the stability of GD. There are three advantages for using MGD over GD and SGD: 1) The batching allows both the efficiency of memory usage and implementations; 2) The model update frequency is higher than GD which allows for a more robust convergence avoiding local minimum; 3) MGD requires less iteration per epoch and provides a more stable update than SGD. For these reason, MGD has been a popular algorithm for learning. However, selecting an appropriate batch size is difficult. Various studies suggest that there is a close link between performance and batch size used in MGD~\cite{ breuel2015effects, keskar2016large, wilson2003general}. 

There are various guidelines for selecting a batch size but have not been completely practical~\cite{bengio2012practical}. Grid search is a popular method but it comes at the expense of search time. There are a small number of adaptive MGD algorithms to replace grid search~\cite{balles2016coupling,byrd2012sample,de2016big,friedlander2012hybrid}. These algorithms increase the batch size gradually according to their own criterion. However, these algorithms are only applicable for convex function and can not be applied to deep learning. For non-convex optimization, it is difficult to determine the optimal batch size for best performance. 

This paper considers a resizable mini-batch gradient descent (RMGD) algorithm based on a multi-armed bandit for achieving best performance in grid search by selecting an appropriate batch size at each epoch with a probability defined as a function of its previous success/failure. At each epoch, RMGD samples a batch size from its probability distribution, then uses the selected batch size for mini-batch gradient descent. After obtaining the validation loss at each epoch, the probability distribution is updated to incorporate the effectiveness of the sampled batch size. The benefit of RMGD is that it avoids the need for cumbersome grid search to achieve best performance and that it is simple enough to apply to any optimization algorithm using MGD. The detailed algorithm of RMGD are described in Section 4, and experiment results are presented in Section 5. \\

\section{Related Works}

There are only a few published results on the topic of batch size. It was empirically shown that SGD converged faster than GD on a large speech recognition database~\cite{wilson2003general}. It was determined that the range of learning rate resulting in low test errors was considerably getting smaller as the batch size increased on convolutional neural networks. Also, it was observed that small batch size yielded the best test error, while large batch size could not yield comparable low error rate~\cite{breuel2015effects}. It was observed that larger batch size are  more liable to converge to a sharp local minimum thus leading to poor generalization~\cite{keskar2016large}. It was found that the learning rate and the batch size controlled the trade-off between the depth and width of the minima in MGD~\cite{jastrzkebski2017three}.

A small number of adaptive MGD algorithms have been proposed. Byrd et al. (2012) introduced a methodology for using varying samples sizes in MGD~\cite{byrd2012sample}. A relatively small batch size is chosen at the start, then the algorithm chooses a larger batch size when the optimization step does not produce improvement in the target objective function. They assumed that using a small batch size allowed rapid progress in the early stages, while a larger batch size yielded high accuracy. However, this assumption did not corresponded with later researches that reported the degradation of performance with large batch size~\cite{breuel2015effects, keskar2016large, mishkin2017systematic}. Another similar adaptive algorithm, which increases the batch size gradually as the iteration proceeded, was done by Friedlander and Schmidt~\cite{friedlander2012hybrid}. The algorithm uses relatively few samples to approximate the gradient, and gradually increase the number of samples with a constant learning rate. It was observed that increasing the batch size is more effective than decaying the learning rate for reducing the number of iterations~\cite{smith2017don}. However, these increasing batch size algorithm has limitation of directional policy. More flexible adaptive algorithm is needed. Balles et al. (2016) proposed a dynamic batch size adaptation algorithm~\cite{balles2016coupling}. It estimates the variance of the stochastic gradients and adapts the batch size to decrease the variance proportionally to the value of the objective function. However, computation cost of this algorithm is expensive and it is not simple enough to apply on large dataset with complex model.\\

\section{Setup}

Let $\mathcal{B} = \{b_k \}_{k=1}^{K}$ be a batch size set and $\bm{\pi} = \{\pi^{k} \}_{k=1}^{K}$ be a probability distribution of batch size where $b_k$, $\pi^k$, and $K$ are the $k^{\mbox{th}}$ batch size, the probability of $b_k$ to be selected, and number of batch sizes respectively. This paper considers algorithm for multi-armed bandit over $\mathcal{B}$ according to Algorithm \ref*{alg:RMGD}. Let $\bm{w}_{\tau} \in \mathcal{W}$ be the model parameters at epoch $\tau$, and $\tilde{\bm{w}}_{t}$ be the temporal parameters at sub iteration $t$. Let $J : \mathcal{W} \rightarrow \mathbb{R}$ be the training loss function and let $\bm{g} = \nabla J(\bm{w})$ be the gradients of training loss function with respect to the model parameters. $\eta_{\tau}$ is the learning rate at epoch $\tau$. Let $\ell : \mathcal{W} \rightarrow \mathbb{R}$ be the validation loss function, and $y^k \in \{0, 1 \}$ be the cost of choosing the batch size $b_k$. In here, $y^k = 0$ if the validation loss decreases by the selected batch size $b_k$ ('well-updating') and $y^k = 1$ otherwise ('misupdating'). The aim of the algorithm is to have low misupdating. \\

\begin{figure*}[t]
	\begin{center}
		\includegraphics[width=0.9\linewidth]{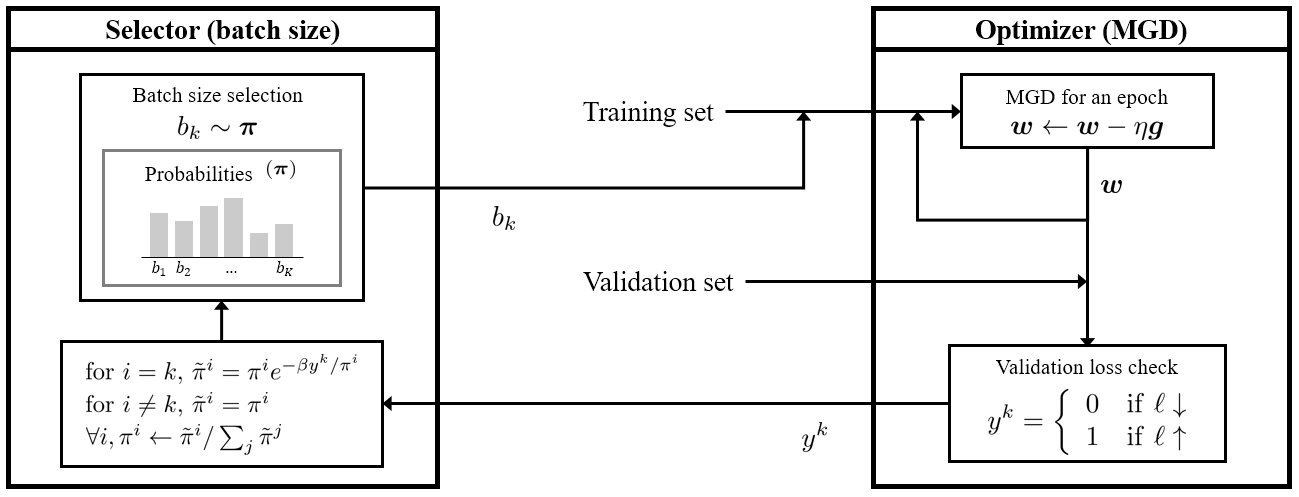}
	\end{center}
	\caption{An overall framework of considered resizable mini-batch gradient descent algorithm (RMGD). The RMGD samples a batch size from a probability distribution, and parameters are updated by mini-batch gradient using the selected batch size. Then the probability distribution is updated by checking the validation loss.}
	\label{fig:framework}
\end{figure*}

\section{Resizable Mini-batch Gradient Descent}

The resizable mini-batch gradient descent (RMGD) sets the batch sizes as multi arms, and at each epoch it samples one of the batch sizes from probability distribution. Then, it suffers a cost of selecting this batch size. Using the cost, probability distribution is updated.\\

\subsection{Algorithms}

The overall framework of the RMGD algorithm is shown in Figure \ref{fig:framework}. The RMGD consists of two components: batch size selector and parameter optimizer. The selector samples a batch size from probability distribution and updates the distribution. The optimizer is usual mini-batch gradient. \\

\textbf{Selector} samples a batch size $b_{k_\tau} \in \mathcal{B}$ from the probability distribution $\bm{\pi}_\tau$ at each epoch $\tau$ where $k_\tau$ is selected index. Here $b_{k}$ is associated with probability  $\pi^{k}$. The selected batch size $b_{k_tau}$ is applied to optimizer for MGD at each epoch, and the selector gets cost $y^{k_\tau}$ from optimizer. Then, the selector updates probabilities,
\begin{align*}
\text{for } i = k_\tau,& \quad \tilde{\pi}^i = \pi_{\tau}^i e^{-\beta y^{k_\tau} / \pi_\tau^i} \nonumber\\
\text{for } i \ne k_\tau,& \quad \tilde{\pi}^i = \pi_{\tau}^i\nonumber\\
\forall i, & \quad \pi_{\tau+1}^i = \tilde{\pi}^i / \sum_j \tilde{\pi}^j
\end{align*}
where $\beta \in (0, 1)$ is positive hyperparameter. When $\tau=0$, $\bm{\pi}_{\tau} = \{1/K, \ldots, 1/K \}$.

\vspace{9px}
\textbf{Optimizer} updates the model parameters $\bm{w}$. For each epoch, temporal parameters $\tilde{\bm{w}}_0$ is set to $\bm{w}_\tau$, and MGD iterates $T = \lceil m / b_{k_\tau} \rceil$\footnote{$\lceil x \rceil$ is the least integer that is greater than or equal to $x$} times using the selected batch size $b_{k_\tau}$ where $m$ is the total number of training samples:
\begin{equation*}
\tilde{\bm{w}}_{t+1} = \tilde{\bm{w}}_{t} - \eta_{\tau} \bm{g}_t, \quad \bm{g}_t = \nabla J (\tilde{\bm{w}}_{t}).
\end{equation*}
After $T$ iterations at epoch $\tau$, the model parameters is updated as $\bm{w}_{\tau+1} = \tilde{\bm{w}}_T$. Then, the optimizer obtains validation loss $\ell$, and outputs cost as follows:
\begin{equation*}
y^{k_\tau} = \left\{ \begin{array}{cl}
0 & \mathrm{if} \ \ell(\bm{w}_{\tau+1}) < \ell(\bm{w}_{\tau}) \\ 
1 & \mathrm{otherwise}
\end{array} \right..
\end{equation*}

\begin{algorithm} [!h]
	\scalebox{0.95}{    
		\begin{minipage}[h!]{1\linewidth}
			\caption{Resizable Mini-batch Gradient Descent}  
			\label{alg:RMGD}
			\textbf{Input}: \\
			$\mathcal{B} = \{b_k\}_{k=1}^K$ : Set of batch sizes \\
			$\bm{\pi}_0 = \{1/K, \ldots, 1/K \} $ : Prior probability distribution \\
			
			\textbf{Procedure:}
			\begin{algorithmic} [1] 
				\STATE Initialize model parameters $\bm{w}_0 $
				\STATE \textbf{for} epoch $\tau = 0, 1, 2, \ldots$				
				\STATE \quad Select batch size $b_{k_\tau} \in \mathcal{B}$ from $\bm{\pi}_\tau$
				\STATE \quad Set temporal parameters $\tilde{\bm{w}}_0 = \bm{w}_\tau$
				\STATE \quad \textbf{for} $ t = 0, 1, \ldots, T-1$ where $T = \lceil m / b_{k_\tau} \rceil$
				\STATE \quad \quad Compute gradient  $\bm{g}_t = \nabla J(\tilde{\bm{w}}_{t}) $
				\STATE \quad \quad Update $\tilde{\bm{w}}_{t+1} = \tilde{\bm{w}}_{t} - \eta_\tau \bm{g}_t$
				\STATE \quad \textbf{end for}
				\STATE \quad Update $\bm{w}_{\tau+1} = \tilde{\bm{w}}_T$
				\STATE \quad Observe validation loss $\ell(\bm{w}_{\tau+1})$
				\STATE \quad \textbf{if} $\ell(\bm{w}_{\tau+1}) < \ell(\bm{w}_{\tau})$
				\STATE \quad \quad Get cost $y^{k_\tau} = 0$
				\STATE \quad \textbf{else}
				\STATE \quad \quad Get cost $y^{k_\tau} = 1$
				\STATE \quad \textbf{end if}
				\STATE \quad \textbf{for} $i = 1, 2, \ldots, K$
				\STATE \quad \quad \textbf{if} $i = k_\tau$
				\STATE \quad \quad \quad Set temporal probability $\tilde{\pi}^i = \pi_{\tau}^i e^{-\beta y^{k_\tau} / \pi_\tau^i}$
				\STATE \quad \quad \textbf{else}
				\STATE \quad \quad \quad Set temporal probability $\tilde{\pi}^i = \pi_{\tau}^i$
				\STATE \quad \quad \textbf{end if}
				\STATE \quad \textbf{end for}
				\STATE \quad Update $\forall i \in [K]$, $\pi_{\tau+1}^{i} = \tilde{\pi}^i / \sum_j \tilde{\pi}^j$
				\STATE \textbf{end for}
			\end{algorithmic}
		\end{minipage}
	}
\end{algorithm}

\begin{figure}[t]
	\begin{center}
		\includegraphics[width=1\linewidth]{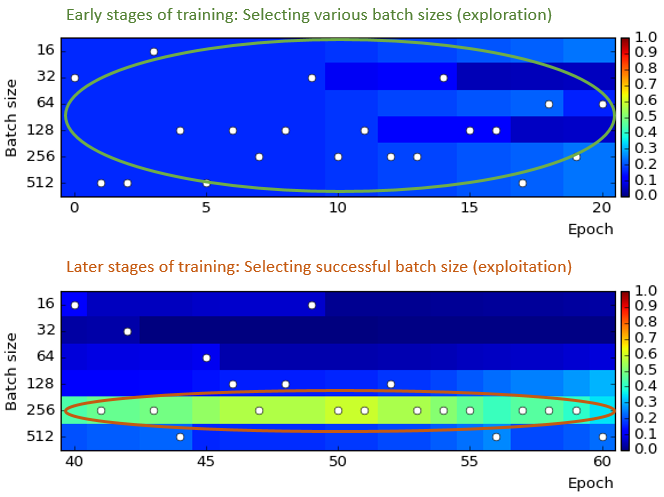}
	\end{center}
	\caption{The probability distribution vs epoch using the RMGD. (top) The early stages of the training. (bottom) The later stages of the training. The white dot represents the selected batch size at each epoch. In the early stages of the training, RMGD updates the probabilities to search various batch sizes (exploration), and in the later stages, RMGD increases the probability of successful batch size (exploitation).}
	\label{fig:mechanism}
\end{figure}

The RMGD samples an appropriate batch size from a probability distribution at each epoch. This probability distribution encourages exploration of different batch size and then later exploits batch size with history of success, which means decreasing validation loss. Figure \ref{fig:mechanism} shows an example of training progress of RMGD. The figure represents the probability distribution with respect to the epoch. The white dot represents the selected batch size at each epoch. In the early stages of the training, commonly, all batch sizes tend to decrease the validation loss, and it makes $\bm{\pi}$ maintain almost uniform distribution. Thus, all batch size have similar probability to be sampled (\emph{exploration}). In the later stages of the training, the probability distribution varies based on success and failure. Thus, better performing batch size gets higher probability to be sampled than others (\emph{exploitation}). In this case, 256 is the best performing batch size.\\

\subsection{Regret Bound}

The analysis of the regret bound of the RMGD is a reconstruction of the multi-armed bandit section of the Shalev-Shwartz's paper~\cite{shalev2012online} to match the RMGD setting. In the RMGD algorithm, there are $K$ batch sizes as multi arms with the probability distribution $\bm{\pi} \in S$, and at each epoch the algorithm should select one of the batch sizes $b_{k_\tau}$. Then it receives a cost of selecting this arm, $y_\tau^{k_\tau} \in \{0, 1 \}$ by testing the validation loss $\ell$. The vector $\bm{y}_\tau \in \{0, 1 \}^K$ represents the selecting cost for each batch size. The goal of this algorithm is to have low regret for not selecting the best performing batch size.
\begin{equation*}
\mathrm{Regret}_\mathcal{T} (S) = \mathbb{E} \left[\sum_{\tau=1}^{\mathcal{T}} y_{\tau}^{k_\tau} \right] - \min_i \sum_{\tau=1}^{\mathcal{T}} y_{\tau}^i
\end{equation*}
where the expectation is over the algorithm's randomness of batch size selection. To analyze the regret bound of the RMGD, online mirror descent (OMD) with estimated gradients algorithm is applied. For OMD setting, let $S$ be the probability simplex and the selecting loss functions be $f_\tau(\bm{\pi}) = \langle \bm{\pi}, \bm{y}_\tau \rangle$. The algorithm select a batch size with probability $\mathbb{P}[b_{k_\tau}] = \pi_{\tau}^{k_\tau}$ and therefore $f_\tau (\bm{\pi}_\tau)$ is the expected cost of the selected batch size at epoch $\tau$. The gradient of the selecting loss function is $\bm{y}_\tau$. However, only one element $y_\tau^{k_\tau}$ is known at each epoch. To estimate gradient, random vector $\bm{z}_\tau$ is defined as follows:
\begin{equation*}
z_\tau^{i} = \left\{ \begin{array}{cl}
y_\tau^i / \pi_{\tau}^i & \mathrm{if} \ i = k_\tau \\ 
0 & \mathrm{otherwise}
\end{array} \right.
\end{equation*}
and expectation of $\bm{z}_\tau$ satisfies,
\begin{equation*}
\mathbb{E}[\bm{z}_\tau | \bm{z}_{\tau-1}, \ldots, \bm{z}_0] = \sum_{i=1}^K \mathbb{P} [b_{k_\tau}] z_\tau^i = \pi_{\tau}^{k_\tau} \frac{y_\tau^{k_\tau}}{\pi_{\tau}^{k_\tau}} = y_\tau^{k_\tau}.
\end{equation*}
The probability distribution $\bm{\pi}_\tau$ is updated by the rule of the normalized exponentiated gradient (normalized-EG) algorithm described in Algorithm \ref*{alg:RMGD}. The selecting loss function is linear and it is satisfied that $\forall \tau, i$ we have $\beta z_\tau^i \ge -1$. Then,
\begin{equation*}
\sum_{\tau=1}^{\mathcal{T}} \langle \bm{\pi}_\tau - \bm{\pi}^* , \bm{z}_\tau \rangle \le \frac{\log(K)}{\beta} + \beta \sum_{\tau=1}^\mathcal{T} \sum_{i=1}^K \pi_{\tau}^i (z_\tau^i)^2
\end{equation*}
where $\bm{\pi}^* \in S$ is a fixed vector which minimizes the cumulative selecting loss,
\begin{equation*}
\bm{\pi}^* = \argmin_{\bm{\pi} \in S} \sum_{\tau=1}^\mathcal{T} f_\tau (\bm{\pi}).
\end{equation*}
Since $f_\tau$ is convex and $\bm{z}_\tau$ is estimated gradients for all $\tau$,
\begin{equation*}
\mathbb{E} \left[ \sum_{\tau=1}^\mathcal{T} (f_\tau (\bm{\pi}_\tau) - f_\tau (\bm{\pi}^*)) \right] \le \frac{\log(K)}{\beta} + \beta \sum_{\tau=1}^\mathcal{T} \mathbb{E} \left[ \sum_{i=1}^K \pi_{\tau}^i (z_\tau^i)^2 \right].
\end{equation*}

The last term is bounded as follows:
\begin{eqnarray*}
	\mathbb{E} \left[ \sum_{i=1}^K \pi_{\tau}^i (z_\tau^i)^2 \right] &=& \sum_{j=1}^K \mathbb{P}[k_\tau = j] \sum_{i=1}^K \pi_{\tau}^i (z_\tau^i)^2 \\
	&=& \sum_{j=1}^K (\pi_{\tau}^j)^2 (y_\tau^j / \pi_{\tau}^j)^2 \\
	&=& \sum_{j=1}^K (y_\tau^j)^2 \le K.
\end{eqnarray*}

Therefore, the regret of the RMGD is bounded,
\begin{equation*}
\mathbb{E} \left[\sum_{\tau=1}^{\mathcal{T}} y_{\tau}^{k_\tau} \right] - \min_i \sum_{\tau=1}^{\mathcal{T}} y_{\tau}^i \le \frac{\log K}{\beta} + \beta K \mathcal{T}.
\end{equation*}

In particular, setting $\beta = \sqrt{\log (K) / (K \mathcal{T})}$, the regret is bounded by $2 \sqrt{K \log (K) \mathcal{T}}$, which is sublinear with $\mathcal{T}$.\\

\begin{figure*}[t]
	\begin{center}
		\includegraphics[width=1\linewidth]{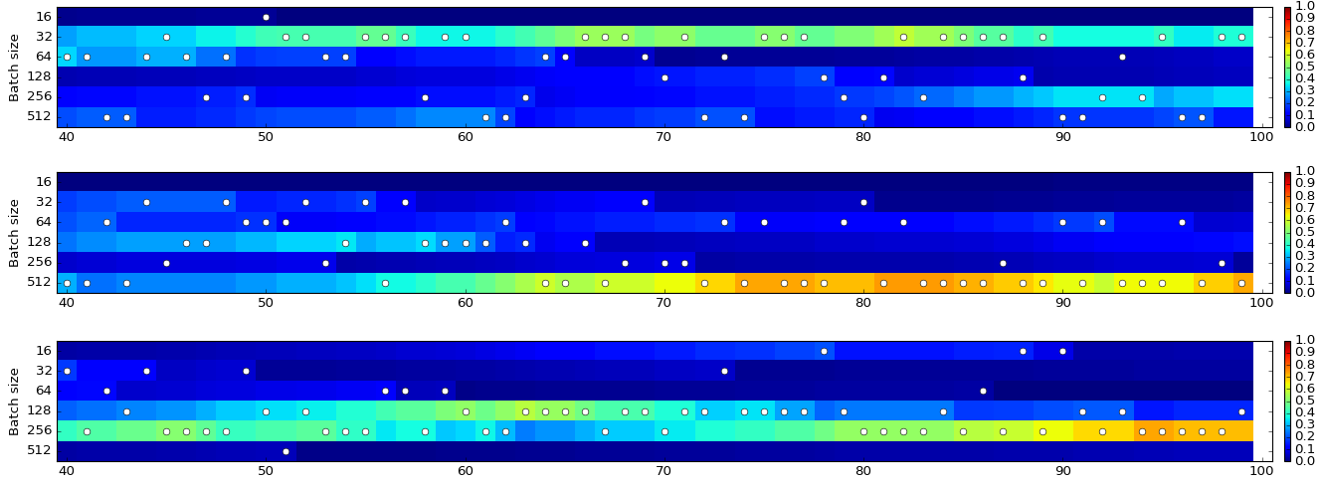}
	\end{center}
	\vspace{-10px}
	\caption{The probability distribution and selected batch size. The white dot is selected batch size at epoch. (top) The case that small batch size performs better. (middle) The case that large batch size performs better. (bottom) The case that best performing batch size varies.}
	\label{fig:res_ps}
\end{figure*}

\section{Experiments}

This section describes the dataset, model settings, and various experimental results. Experimental results comparing the performance of the RMGD with that of the MGD are presented. \\

\subsection{Dataset}

\textbf{MNIST} is a dataset of handwritten digits that is commonly used for image classification. Each sample is a black and white image and $28 \times 28$ in size. The MNIST is split into three parts: 55,000 samples for training, 5,000 samples for validation, and 10,000 samples for test. \\

\textbf{CIFAR10} consists of 60,000 $32 \times 32$ color images in 10 classes (airplane, automobile, bird, cat, deer, dog, frog, horse, ship, and truck), with 6,000 images per class. The CIFAR10 is split into three parts: 45,000 samples for training, 5,000 samples for validation, and 10,000 samples for test. \\

\subsection{Settings}
In the experiments, simple convolutional neural networks (CNN) is used for MNIST and `All-CNN-C'~\cite{springenberg2014striving} is used for CIFAR10. The simple CNN consists of two convolution layers with $5 \times 5$ filter and $1 \times 1$ stride, two max pooling layers with $2 \times 2$ kernel and $2 \times 2$ stride, single fully-connected layer, and softmax classifier.
Description of the 'All-CNN-C' is provided in Table \ref{model_allconv}. For MNIST, AdamOptimizer with $\eta = 10^{-4}$ and AdagradOptimizer with $\eta = 0.1$ are used as optimizer, and $\mathcal{B} = \{16, 32, 64, 128, 256, 512 \}$. The model is trained for a total of 100 epochs. For `All-CNN-C', MomentumOptimizer with fixed momentum of 0.9 is used as optimizer. The learning rate $\eta^k$ is scaled up proportionately to the batch size ($\eta^k = 0.05 * b_k / 256$, ~\cite{breuel2015effects}) and decayed by a schedule $S = [200, 250, 300]$ in which $\eta^k$ is multiplied by a fixed multiplier of 0.1 after 200, 250, and 300 epochs respectively. The model is trained for a total of 350 epochs. Dropout is applied to the input image as well as after each convolution layer with stride 2. The dropout probabilities are 20\% for dropping out inputs and 50\% otherwise. The model is regularized with weight decay $\lambda = 0.001$, and  $\mathcal{B} = \{16, 32, 62, 128, 256\}$. For all experiments, rectified linear unit (ReLU) is used as activation function. For RMGD, $\beta$ is set to $\sqrt{\log(6)/(6*100)} \approx 0.055$ and $\sqrt{\log(5)/(5*350)} \approx 0.030$ for MNIST and CIFAR10 respectively. \\

\begin{table}[!t]
	\centering
	\caption{Architecture of the All-CNN-C for CIFAR10}
	\label{model_allconv}
	\begin{tabular}{c|c}
		\hline
		Layer      & Layer description  \\ \hline
		input      & Input 32 $\times$ 32 RGB image \\ 
		conv1      & 3 $\times$ 3 conv. 96 ReLU, stride 1, dropout 0.2 \\ 
		conv2      & 3 $\times$ 3 conv. 96 ReLU, stride 1 \\
		conv3      & 3 $\times$ 3 conv. 96 ReLU, stride 2 \\
		conv4      & 3 $\times$ 3 conv. 192 ReLU, stride 1, dropout 0.5 \\
		conv5      & 3 $\times$ 3 conv. 192 ReLU, stride 1 \\
		conv6      & 3 $\times$ 3 conv. 192 ReLU, stride 2 \\
		conv7      & 3 $\times$ 3 conv. 192 ReLU, stride 1, dropout 0.5 \\
		conv8      & 1 $\times$ 1 conv. 192 ReLU, stride 1 \\
		conv9      & 1 $\times$ 1 conv. 10 ReLU, stride 1 \\
		pool       & averaging over 6 $\times$ 6 spatial dimensions \\
		softmax    & 10-way softmax \\ \hline
	\end{tabular}
\end{table}

\begin{figure*}[t]
	\begin{center}
		\includegraphics[width=1\linewidth]{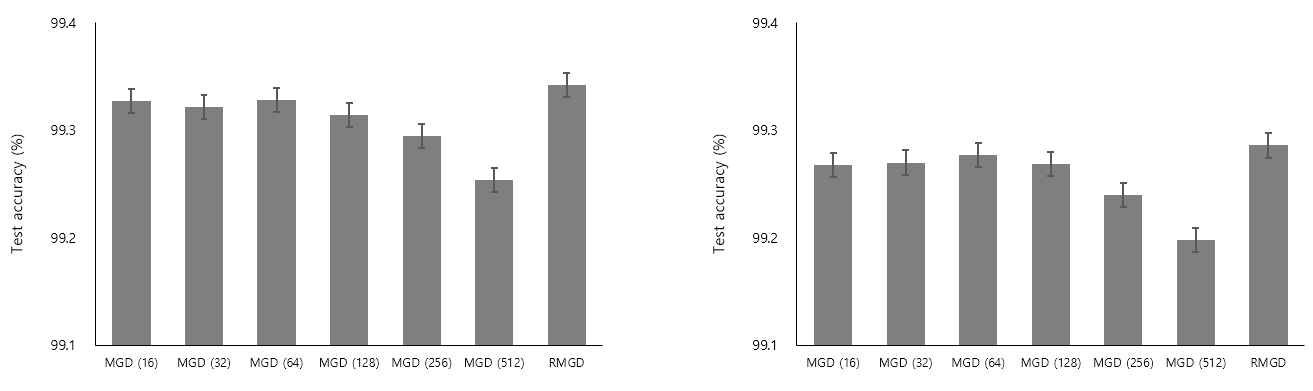}
	\end{center}
	\caption{The results of test accuracy for the MNIST dataset. The number in parenthesis next to MGD represents the batch size used in the MGD. (left) The test accuracy of 100 times repeated experiments with AdamOptimizer. The error bar is standard error. (right) The test accuracy of 100 times repeated experiments with AdagradOptimizer. In both cases, the RMGD outperforms all fixed MGD algorithm.}
	\label{fig:res_accuracy}
\end{figure*}

\begin{table*}[!h]
	\centering
	\caption{Iterations and real time for training, and test accuracy of MNIST classification with AdamOptimizer.}
	\label{table:Adam}
	\begin{tabular}{|c|c|c|c|c|c|}
		\hline
		\quad \quad Algorithms \quad \quad &\ \quad Iterations \quad \ & \quad Real time (sec) \quad & \multicolumn{3}{c|}{Test accuracy (\%)} \\ \cline{4-6} 
		&            &                & \quad \quad Mean $\pm$ SD \quad \quad &\quad \quad  Max \quad \quad & \quad \quad Min \quad \quad  \\ \hline
		MGD (16)     & 343,800  & 1,221.54 $\pm$ 36.00 & 99.327 $\pm$ 0.064 & 99.480 & 99.140 \\ \hline
		MGD (32)     & 171,900  & 697.82 $\pm$ 19.70 & 99.322 $\pm$ 0.060 & 99.500 & 99.150 \\ \hline
		MGD (64)     & 86,000  &  379.14 $\pm$ 11.32 & 99.328 $\pm$ 0.058 & 99.460 & 99.170 \\ \hline
		MGD (128)    & 43,000 &  262.33 $\pm$ 2.34 & 99.314 $\pm$ 0.056 & 99.440 & 99.170 \\ \hline
		MGD (256)    & 21,500 &  208.13 $\pm$ 2.20 & 99.295 $\pm$ 0.059 & 99.470 & 99.170 \\ \hline
		MGD (512)    & 10,800 &  180.06 $\pm$ 0.37 & 99.254 $\pm$ 0.054 & 99.430 & 99.110 \\ \hline 
		MGD (total)  & 677,000             &  2,949.02            &                    &        &        \\ \hline 
		RMGD     & 68,309 $\pm$  8,900 &  333.73 $\pm$ 25.38 & \textbf{99.342} $\pm$ \textbf{0.064} & 99.480 & 99.110 \\ \hline 
	\end{tabular}
\end{table*}

\begin{table*}[!h]
	\centering
	\caption{Iterations and real time for training, and test accuracy of MNIST classification with AdagradOptimizer.}
	\label{table:Adagrad}
	\begin{tabular}{|c|c|c|c|c|c|}
		\hline
		\quad \quad Algorithms \quad \quad &\ \quad Iterations \quad \ & \quad Real time (sec) \quad & \multicolumn{3}{c|}{Test accuracy (\%)} \\ \cline{4-6} 
		&            &                & \quad \quad Mean $\pm$ SD \quad \quad &\quad \quad  Max \quad \quad & \quad \quad Min \quad \quad  \\ \hline
		MGD (16)     & 343,800  & 1,160.87 $\pm$ 22.34 & 99.268 $\pm$ 0.090 & 99.430 & 98.920 \\ \hline
		MGD (32)     & 171,900  & 640.68 $\pm$ 15.53 & 99.270 $\pm$ 0.070 & 99.410 & 99.050 \\ \hline
		MGD (64)     & 86,000  &  367.40 $\pm$ 12.63 & 99.277 $\pm$ 0.077 & 99.440 & 99.110 \\ \hline
		MGD (128)    & 43,000 &  262.48 $\pm$ 1.37 & 99.269 $\pm$ 0.069 & 99.410 & 99.080 \\ \hline
		MGD (256)    & 21,500 &  195.60 $\pm$ 2.00 & 99.240 $\pm$ 0.072 & 99.390 & 99.030 \\ \hline
		MGD (512)    & 10,800 &  170.31 $\pm$ 1.41 & 99.198 $\pm$ 0.085 & 99.390 & 98.810 \\ \hline 
		MGD (total)  & 677,000             &  2,797.34            &                    &        &        \\ \hline 
		RMGD     & 68,159 $\pm$  8,447 &  323.33 $\pm$ 23.57 & \textbf{99.286} $\pm$ \textbf{0.088} & 99.490 & 98.900 \\ \hline 
	\end{tabular}
\end{table*}

\begin{figure*}[t]
	\begin{center}
		\includegraphics[width=0.9\linewidth]{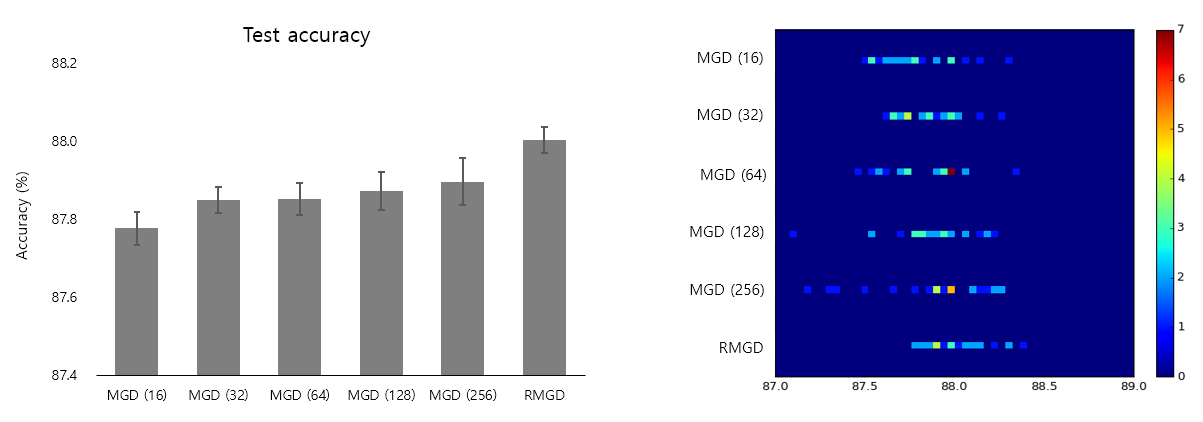}
	\end{center}
	\caption{The results of test accuracy for the CIFAR10. (left) The mean test accuracy of 25 times repeated experiments for each algorithm. The error bar is standard error. (right) Histogram of test accuracies for each algorithm.}
	\label{fig:result_allconv}
\end{figure*}

\begin{table*}[h]
	\centering
	\caption{Iterations and relative time for training, and test accuracy for CIFAR10 with MomentumOptimizer. Each experiment was repeated 25 times.}
	\label{table:cifar10_allconv}
	\begin{tabular}{|c|c|c|c|c|c|}
		\hline
		\quad \quad Algorithms \quad \quad &\ \quad Iterations \quad \ & \quad Real time (sec) \quad & \multicolumn{3}{c|}{Test accuracy (\%)} \\ \cline{4-6} 
		&            &                & \quad \quad Mean $\pm$ SD \quad \quad &\quad \quad  Max \quad \quad & \quad \quad Min \quad \quad  \\ \hline
		MGD (16)     & 1,072,050 & 10,085.26 $\pm$ 216.48 & 87.778 $\pm$ 0.207 & 88.290 & 87.480 \\ \hline
		MGD (32)     & 536,200 & 7,643.93 $\pm$ 459.95 & 87.851 $\pm$ 0.160 & 88.250 & 87.630 \\ \hline
		MGD (64)     & 268,100 & 6,160.16 $\pm$  68.54 & 87.853 $\pm$ 0.202 & 88.330 & 87.450 \\ \hline
		MGD (128)    & 134,050 & 5,675.15 $\pm$ 181.80 & 87.873 $\pm$ 0.234 & 88.210 & 87.090 \\ \hline
		MGD (256)    &  67,200 & 5,466.79 $\pm$ 402.20 & 87.897 $\pm$ 0.293 & 88.260 & 87.170 \\ \hline
		MGD (total)  & 2,077,600  & 35,031.29 &                    &        &        \\ \hline 
		RMGD         & 463,629 $\pm$ 48,692  & 7,592.43 $\pm$ 403.65 & \textbf{88.004} $\pm$ \textbf{0.167} & 88.380 & 87.780 \\ \hline
	\end{tabular}
\end{table*}

\subsection{Results}
The validity of the RMGD was assessed by performing image classification on the MNIST dataset using AdamOptimizer and AdagradOptimizer as optimizer. The experiments were repeated 100 times for each algorithm and each optimizer, then the results were analyzed for significance. Figure \ref{fig:res_ps} shows the probability distribution and the selected batch size with respect to epoch during training for the RMGD. The white dot represents the batch size selected at each epoch. The top figure is the case that small batch size (32) performs better. After epoch 50, batch size 32 gets high probability and is selected more than others. It means that batch size 32 has less misupdating in this case. The gradually increasing batch size algorithm may not perform well in this case. The middle figure is the case that large batch size (512) performs better. After epoch 60, batch size 512 gets high probability and selected more than others. The bottom figure is the case that best performing batch size varies. During epoch from 40 to 55, batch size 256 performs better, and better performing batch size changes to 128 during epoch from 60 to 70, then better performing batch size comes back to 256 after epoch 80. In the results, any batch size can be a successful batch size in the later stages without any particular order. The RMGD is more flexible for such situation than the MGD or directional adaptive MGD such as gradually increasing batch size algorithm.

Figure \ref{fig:res_accuracy} shows the test accuracy for each algorithm. The error bar is standard error. The number in parenthesis next to MGD represents the batch size used in the MGD. The left figure is the test accuracy with AdamOptimizer. The right figure is the test accuracy with AdagradOptimizer. Among the MGD algorithms, relatively small batch sizes (16 - 64) lead to higher performance than large batch sizes (128 - 512) and batch size 64 achieves the best performance. These results correspond with other studies~\cite{breuel2015effects, keskar2016large, mishkin2017systematic}. The RMGD outperforms all fixed MGD algorithm in both case. Although the performance of RMGD is not significantly increased compared to the MGD, the purpose of this algorithm is not to improve performance, but to ensure that the best performance is achieved without performing a grid search on the batch size. Rather, the improved performance of the RMGD is a surprising result. Therefore, the RMGD is said to be valid. 

Table \ref{table:Adam} and \ref{table:Adagrad} present iterations and real time for training, mean, maximum, and minimum of test accuracies for each algorithm with AdamOptimizer and AdagradOptimizer respectively. The MGD (total) is the summation of the iterations and real time of whole MGDs for grid search. For the MGD, there is trade-off between the performance and the training time. The RMGD outperforms best performance of the MGD faster than best performing MGD in both cases. Furthermore, it is 8 times faster than grid search in both cases. In the results, the RMGD is effective regardless of the optimizer. \\

The CIFAR10 dataset was, also, used to assess effectiveness of the RMGD. The experiments were repeated 25 times for each algorithm. In this experiment, all images are whitened and contrast normalized before being input to the network. Figure~\ref{fig:result_allconv} shows the test accuracy for each algorithm. The left figure represents the mean test accuracy with standard error. In contrast to the MNIST results, relatively large batch sizes (128 - 256) lead to higher performance than small batch sizes (16 - 64) and batch size 256 achieves the best performance. The results of MNIST and CIFAR10 indicate that it is difficult to know which batch size is optimal before performing a grid search. Meanwhile, the RMGD has again exceeded the best performance of fixed MGD. The right figure represents the histogram of test accuracies for each algorithm. The test accuracies of the RMGD are distributed at high values. 

Table~\ref{table:cifar10_allconv} presents the detailed results on CIFAR10 dataset. The RMGD is a little slower than single best performing MGD (256), however, it was much faster than grid search -about 4.6 times faster. Therefore, this results, also, show the effectiveness of the RMGD. \\

The t-test was performed to validate the significance of the RMGD and p-values\footnote{P-value level: $*(<0.05)$, $**(<0.01)$, $***(<0.001)$} of t-test are presented in Table \ref{table:ttest}. This is a one-sided t test for the null hypothesis that the expected value (mean) of the RMGD's test accuracies is smaller than the best mean test accuracy of fixed MGD. The RMGD significantly achieves the best performance of the grid search using AdamOptimizer on MNIST and MomentumOptimizer on CIFAR10: p-values obtained are $0.019^*$ and $0.002^{**}$ respectively. Also, it can be said that the RMGD achieves the best performance of the grid search with probability of 84.5\% using AdagradOptimizer on MNIST, which p-value is $0.155$. Therefore, the RMGD achieves the best performance of the grid search with high probability and requires much shorter amount of time than the grid search regardless of dataset, model, or optimizer. \\

\begin{table}[t]
	\centering
	\caption{P-values by one-sided t-test for the mean of the RMGD's test accuracy against to the best mean test accuracy of fixed MGD.}
	\label{table:ttest}
	\begin{tabular}{|c|c|}
		\hline
		&\quad  p-value \quad  \\ \hline
		MNIST (AdamOptimizer)  &  $\mathbf{0.019^*}$  \\ \hline
		MNIST (AdagradOptimizer)  & $0.155$    \\ \hline
		\quad CIFAR10 (MomentumOptimizer) \quad & \quad \  $\mathbf{0.002^{**}}$  \quad \   \\ \hline
	\end{tabular}
\end{table}


\section*{Conclusion}
Selecting batch size affects the model quality and training efficiency, and determining the appropriate batch size when performing mini-batch gradient descent is always time consuming and requires considerable resources as it often relies on grid search. This paper considers a resizable mini-batch gradient descent (RMGD) algorithm based on a multi-armed bandit for achieving best performance in grid search by selecting an appropriate batch size at each epoch with a probability defined as a function of its previous success/failure. This probability encourages exploration of different batch size and then later exploitation of batch size with history of success. The goal of this algorithm is not to achieve state-of-the-art accuracy but rather to select appropriate batch size which leads low misupdating and performs better. The RMGD essentially assists the learning process to explore the possible domain of the batch size and exploit successful batch size. The benefit of RMGD is that it avoids the need for cumbersome grid search to achieve best performance and that it is simple enough to apply to various field of machine learning including deep learning using MGD. Experimental results show that the RMGD achieves the best grid search performance with \textit{high probability} on MNIST and CIFAR10.
Furthermore, it, obviously, attains this performance in a shorter amount of time than the grid search. In conclusion, the RMGD is effective and flexible mini-batch gradient descent algorithm.\\

{\small
\bibliographystyle{ieee}
\bibliography{egbib}
}

\end{document}